\documentclass{article}

\PassOptionsToPackage{numbers, compress}{natbib}
\usepackage[preprint]{neurips_2021}

\usepackage[utf8]{inputenc} % allow utf-8 input
\usepackage[T1]{fontenc}    % use 8-bit T1 fonts
\usepackage{hyperref}       % hyperlinks
\usepackage{url}            % simple URL typesetting
\usepackage{booktabs}       % professional-quality tables
\usepackage{amsfonts}       % blackboard math symbols
\usepackage{nicefrac}       % compact symbols for 1/2, etc.
\usepackage{microtype}      % microtypography
\usepackage{xcolor}         % colors
\usepackage{graphicx}
\usepackage{subcaption}
\usepackage{array}
\usepackage{lipsum}
\usepackage{amsmath,amssymb}
\usepackage{multirow}
\usepackage{multicol}
\definecolor{carmine}{rgb}{0.59, 0.0, 0.09}
\definecolor{darkmidnightblue}{rgb}{0.0, 0.2, 0.4}
\newcommand{\RRED}[1]{{\color{carmine}#1}}
\newcommand{\BLUE}[1]{{\color{darkmidnightblue}#1}}
\newcommand{\wh}[1]{{\color{black}#1}}

\title{Revisit Visual Representation in Analytics Taxonomy: A Compression Perspective}
\author{%
  Yueyu Hu, Wenhan Yang, Haofeng Huang, Jiaying Liu\\
  Wangxuan Institute of Computer Technology\\
  Peking University, Beijing, China \\
  \texttt{\{huyy, yangwenhan, huang6013, liujiaying\}@pku.edu.cn} \\
}

\begin{document}

\maketitle

\begin{abstract}
\wh{
% Visual analytics have played an increasingly critical role in Internet of Things and surveillance system, where massive visual signals have to be compressed and fed into machines. But existing image/video compression at the pixel level leads to much redundant representations and low value density for machine analytics. It raises the great challenge to the large-scale visual intelligent analytics system.
Visual analytics have played an increasingly critical role in the Internet of Things, where massive visual signals have to be compressed and fed into machines. But facing such big data and constrained bandwidth capacity, existing image/video compression methods lead to very low-quality representations, while existing feature compression techniques fail to support diversified visual analytics applications/tasks with low-bit-rate representations. In this paper, we raise and study the novel problem of supporting multiple machine vision analytics tasks with the compressed visual representation, namely, the information compression problem in analytics taxonomy. By utilizing the intrinsic transferability among different tasks, our framework successfully constructs compact and expressive representations at low bit-rates to support a diversified set of machine vision tasks, including both high-level semantic-related tasks and mid-level geometry analytic tasks. In order to impose compactness in the representations, we propose a codebook-based hyperprior, which helps map the representation into a low-dimensional manifold. As it well fits the signal structure of the deep visual feature, it facilitates more accurate entropy estimation, and results in higher compression efficiency. With the proposed framework and the codebook-based hyperprior, we further investigate the relationship of different task features owning different levels of abstraction granularity. Experimental results demonstrate that with the proposed scheme, a set of diversified tasks can be supported at a significantly lower bit-rate, compared with existing compression schemes. 
		}
\end{abstract}

\section{Introduction}
	\wh{
	% In the big data era, smart city applications and Internet of Things produce massive visual data everywhere, 
	% %
	% %
	% The new trend is that more and more images and videos are consumed by machine vision analytics systems.
	% %
	% In this case, although video coding systems continuously improve the compression efficiency~\cite{sullivan2012overview,vvc}, it is still far away from the need for fast response of the huge amount of applications by transmitting videos/images and then performing analysis based on the videos/images.
    
  % The bottleneck lies in that the pixel information serving for human vision is much redundant and of low values from the perspective of machine analytics.
  %   %
  % Therefore, it is quite essential and critical to compress visual signals generated everywhere and anytime, in a more compact and efficient way to serve the massive applications, \textit{i.e.} for machines. In general, the desirable \textit{visual information compression for machine analytics} prefers the following properties:
  In the big data era, the Internet of Things generates massive and heterogenous visual data. The data volume is too large that conventional paradigm, \textit{i.e.} compressing videos and analyzing in the cloud, leads to severe compression artifacts and largely degrades the usability and robustness of visual analytics systems. Directly compressing the visual features is more feasible. However, existing techniques cannot support various tasks with a unified representation and transmitting multiple features also consumes large volume bit-rates when dealing with tremendous visual data. Ideally, the desirable visual information compression for machine analytics prefers the following properties:

	\noindent \textbf{Compactness.} Image/video pixels carry much redundant information, some of which is unnecessary for machine vision analytics. 
	The compression scheme should only select the most valuable information for analytics to achieve compactness in the representation, in order to improve the efficiency of the whole processing system.
	
	\noindent \textbf{Versatility.} Due to the diversity of applications, such a scheme has to work with a wide range of analytics tasks/applications. 
	Some tasks handle object/background semantics, \textit{e.g.} object recognition~\cite{russakovsky2015imagenet} and semantic segmentation~\cite{lin2014microsoft}. 
	Other tasks may focus on the geometry of objects/background, including 3D and 2D vision analytics, \textit{e.g.} surface normal estimation~\cite{eigen2015predicting}, depth estimation~\cite{silberman2012indoor}, and edge extraction~\cite{dollar2014fast}.
	
	\noindent \textbf{Scalability.} Constraints on bit-rates and requirements of precision vary among different application scenarios.
	A practical vision analytics system should be flexible to support tasks that require more abundant information when more bit-rates can be provided. 
	Meanwhile, it should be possible to enforce the compactness of the compressed feature when the constraint on the bit-rate is tight.
	
	Some works explore the joint optimization of visual analytics and compression, \textit{e.g.} semantic guided bit allocation~\cite{choi2018objectcompression,he2019beyondcoding}, analytics with image bit-streams~\cite{torfason2018towards}.
  % and joint optimization of image compression with machine vision tasks~\cite{chamain2021end}. 
	%
	Recently, the video coding for machine~(VCM) paradigm~\cite{duan2020video} attempts to bridge the gap between feature coding for machine vision and video compression by typical geometric visual descriptors, \textit{e.g.} key points~\cite{xia2020emerging} and edges~\cite{yang2021towards}. However, these schemes still rely on reconstructed video representation at the low bit-rate, and thus lack versatility and scalability. Meanwhile, there are works on compressing highly abstract visual features~\cite{lowe1999object,duan2015overview,duan2018compact} and deep features~\cite{chen2019toward,cohen2020lightweight}.
  %  These descriptors are utilized to represent images for compression, and generative models are employed to reconstruct images from these descriptors.
	%
	% As these schemes are basically based on the pixel-level representation with low bit-rates, 
	%
	% they fall short of supporting various high-level machine vision tasks with high compression efficiency.
	%
	% Thus, they are still less satisfactory from the perspectives of compactness, versatility, and flexibility.
	%
	% Meanwhile, there are works on compressing highly abstract visual features for machine vision analytics, including the typical Scale-Invariant Feature Transform (SIFT)~\cite{lowe1999object}, Compact
	% descriptors for visual search (CDVS)~\cite{duan2015overview,duan2018compact} and recent neural network~(NN)-generated features~\cite{chen2019toward,cohen2020lightweight}. 
	%
% 	As the learned features are more flexible, we can easily control its functionality, information volume and bit-rate if we design the network architectures and train them in a proper way.
	%
	It has been proposed in \cite{singh2020end} and \cite{dubois2021lossy} to jointly optimize the feature extraction and compression for visual classification. But very deep features without spatial dimensions are adopted, which results in inability to handle various kinds of tasks.
	%
	% However, these schemes still lack the flexibility to handle spatial geometry-related tasks, as the extracted features are too abstract and location-agnostic, which hinders the reconstruction of this spatial geometry information.
	%
	
	It has been analyzed in \cite{zamir2018taskonomy} that different machine vision tasks have underlying transferability among features extracted by learned neural networks. The result indicates that some visual information can be shared among tasks, and there exists cross-task feature redundancy. Thus, it implies the great potential to collaboratively compress a set of visual tasks. However, it is still unclear about the characteristics of different tasks from the perspective of information entropy, and how to fully investigate their complementarity. It is non-trivial to properly aggregate different granularity features to support a variety of tasks jointly.
  % that different machine vision tasks have underlying connections: feature extracted by a learned neural network for one task might be transferred to handle another task.
	%
	% The result indicates that some visual information can be shared among some different tasks and there is cross-task redundancy in their features.
	%
	% Thus, the joint compression for a set of tasks is potentially beneficial to form a more compact representation.
	%
	% However, it is still unclear about the characteristics of different tasks in information entropy and how they complement each other. It is non-trivial to leverage different feature representations to form an aggregated one to support a set of tasks.
	
	In this work, we explore to address the problem of information compression in analytics taxonomy. 
	We study the signal structure of deep feature representations, and propose a codebook-based hyperprior model
	%to describe the probability distribution and 
	to estimate the information entropy of the general visual representations.
	With the proposed method, we study the rate-distortion characteristics of the representations among different tasks. 
	The study leads to an aggregation transformed compression model that generates a unified representation from multiple representations.
	Such a compression scheme saves more bit-rates than the strategy of compressing for each task independently.
    We further explore the potential of the proposed information compression scheme to support external unseen tasks. Our contributions are summarized as follows,}
	\wh{
	\begin{itemize}
		\item To the best of our knowledge, we are the first to formulate and study the problem of visual data compression in analytics taxonomy, where the compression of the unified feature for both high-level semantic-related tasks and mid-level geometry analytic tasks are investigated.
		\item We propose a codebook-based hyperprior model to compress deep feature representations.
		The proposed scheme employs a novel entropy estimation to well fit the signal structure of deep visual data.
		With the proposed scheme, we minimize the bit-rates but still efficiently support different machine vision tasks.
    %study the minimal bit-rates to fully support different machine vision tasks.
		\item With the compression scheme, we further study the joint compression of visual data for a set of tasks. We show that a set of tasks can be supported by unified compressed representation. We also explore the potential of the compressed representation to support unseen tasks.
	\end{itemize}}

% \section{Related Works}

% \subsection{Learned Image Compression}

% Our work has close connections with learned image compression techniques~\cite{hu2021learning}. 

% \subsection{Visual Feature Compression}
\section{Information Compression in Analytics Taxonomy}

\begin{figure}[t]
  \centering
  
    \begin{subfigure}[t]{1\linewidth}%[width=14cm]
    \centering
    \includegraphics[width=0.75\linewidth]{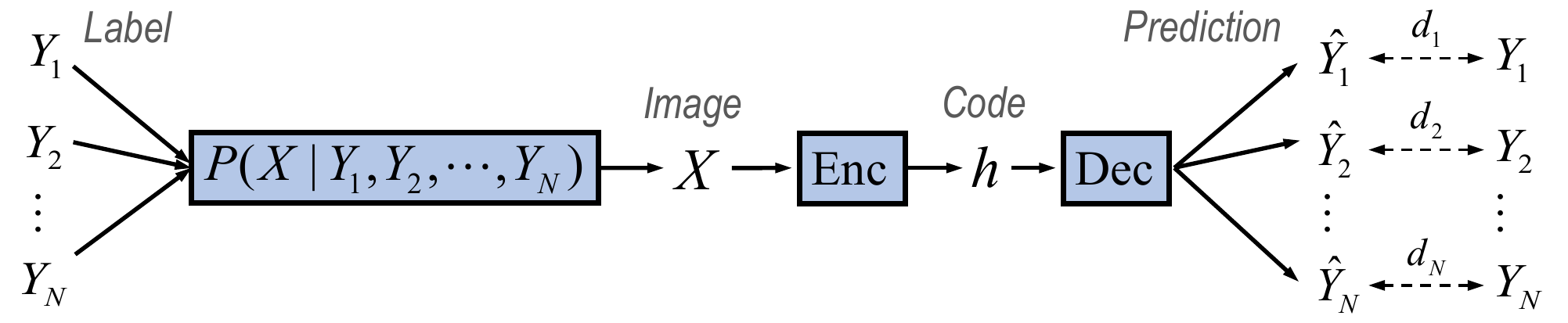}
    \caption{From the remote source coding perspective.}
    \label{fig:indirect}
    \end{subfigure}
    
    % \vspace{1mm}
    
    \begin{subfigure}[t]{1\linewidth}
    \centering
    \includegraphics[width=0.78\linewidth]{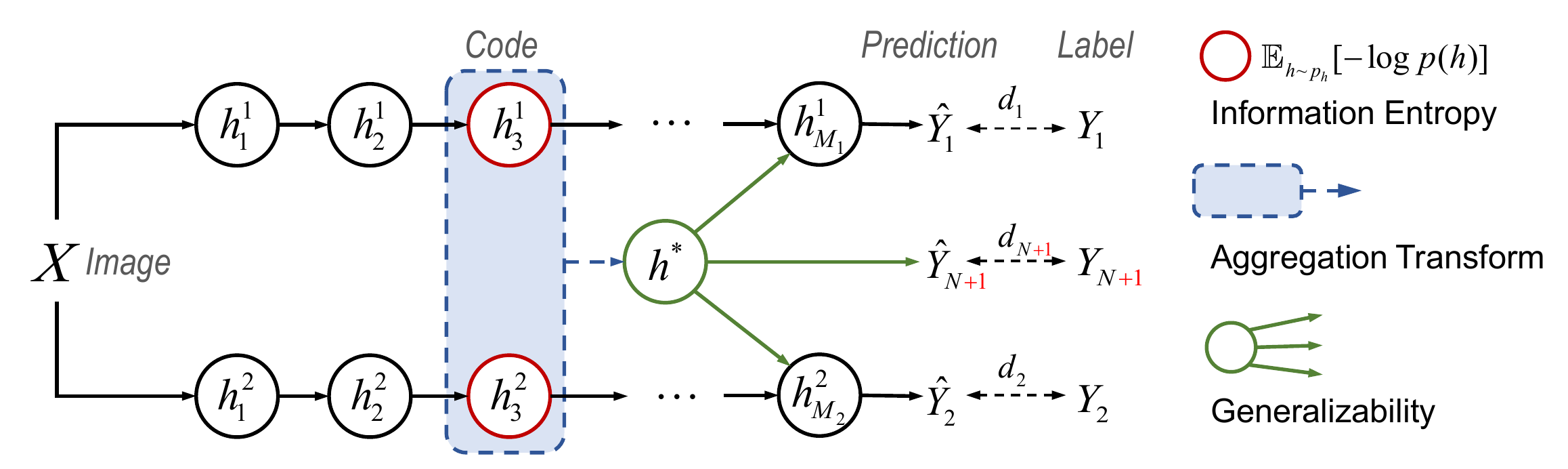}
    \caption{Compression for analytics with knowledgeable neural networks.}
    \label{fig:nn-gen}
    \end{subfigure}

  \caption{
  The problem formulation of the compression for analytics in a collaborative intelligent visual analytics system.
  (a) The goal is to estimate the latent semantic labels from the observed image data, constrained by the bandwidth.
  (b) When the problem is relaxed with the help of pre-trained neural networks, the proposed turns into compact, effective, and general feature extraction, which needs to be optimized in three aspects: information entropy, aggregation transform and generality.
}
\vspace{-2mm}
  \label{fig:frame1}
\end{figure}

\subsection{From the Remote Source Coding Perspective}
\label{sec:fomulate1}

This section aims to formulate the rate-distortion (R-D) optimization problem for information compression in analytics taxonomy.
Compared with the traditional R-D performance of the video compression, the compression for analytics has two distinct differences: 1) the original signal (\textit{i.e.} the ground truth labels) cannot be observed during the encoding process; 2) the compression scheme has to consider aggregated R-D performance measuring a wide range of tasks (including known and unseen) in an emerging collaborative intelligent system. The distinctions lead to the indirect source coding problem~\cite{kipnis2021rate}.

As shown in Fig.~\ref{fig:indirect}, $X$ is the random variable representing the captured image/video by the front-end sensors, which are assumed to be generated by a stochastic process $P_{ {X} | {Y_1, Y_2, ..., Y_N} }$ conditioned on the semantic labels $Y_1, Y_2, ..., Y_N$, where $N$ is the total task number. 
For convenience, we refer $\mathbf{Y}=(Y_1, Y_2, ..., Y_N)$ to the random vector related to a wide range of possible machine vision tasks (known and unseen). It represents the intrinsic semantics.
% This ${Y}$ here is not a random variable representing the simple combination of the semantic labels of several tasks.
% $Y$ corresponds to the joint distribution of the random variables for all labels in a collection, related to a wide range of possible machine vision tasks (existing, known, and unseen). It represents the intrinsic semantics.
%which is a super-set of the semantics of all machine vision task sets.

\wh{
The goal of the compression scheme is to extract semantic predictions $\hat{\mathbf{y}} = (\hat{y}_1, \hat{y}_2, \cdots, \hat{y}_N)$, from the observed image $x \sim P_{X|Y}$, to minimize the distortion under the constraint of bit-rate.
The aggregated distortion $d^{*}\left(\cdot \right)$  \textit{w.r.t.} the semantic prediction $\hat{{\mathbf{y}}}$ and the original semantic label $\mathbf{y} = ({y}_1, {y}_2, \cdots, {y}_N)$ is defined as,
\begin{equation}
  d^{*}\left( \mathbf{y}, \hat{\mathbf{y}} \right) = 
  f \left( d_1 \left( y_1, \hat{y}_1 \right), d_2\left(y_2, \hat{y}_2\right), \cdots, d_N\left( y_N, \hat{y}_N \right) \right),
\end{equation}
where $d_t:\hat{\mathcal{Y}}_t \times \mathcal{Y}_t \rightarrow \mathbb{R}, t\in [1,N]$ is a distortion metric for $\hat{y}_t \in \hat{\mathcal{Y}}_t$ with the ground truth label $y_t \in \mathcal{Y}_t$. $\hat{\mathcal{Y}}_t \times \mathcal{Y}_t$ is the corresponding sample space of $( \hat{{Y}}_t, {Y}_t )$.
$f(\cdot)$ defines the aggregation of different distortion metrics.
%
% Intuitively, the image data ${X}$ as well as $\mathbf{Y}$ matches the form of the popular computer vision image/videos dataset $\{(X, Y)\} \in \mathcal{X}\times\mathcal{Y}$ (the sample space of $(X, Y)$), \textit{e.g.} Taskonomy dataset~\cite{zamir2018taskonomy}, where each $Y_i$ corresponds to a machine vision \textit{task} based on the observed image signal $X$, and altogether forms the analytics taxonomy.
}

% \wh{As some info  rmation about $Y$ is unknown, such as the related distortion metrics $d_i(\cdot)$ of some tasks, as well as the aggregation function when designing the related codecs,}
As $\mathbf{Y}$ cannot be observed when conducting analytics on $X$, \wh{we can assume a Markov chain $\mathbf{Y} \rightarrow X \rightarrow \hat{\mathbf{Y}}$~\cite{tishby2015deep,kipnis2017coding}, namely predicting $\hat{\mathbf{y}}$ from $x$ is not correlated with $\mathbf{y}$. The goal of achieving the optimal analytics performance can be formulated as the following optimization problem,
% In this case, we can reach the R-D cost in the following constrained optimization problem:
\begin{equation}
\begin{split}
  \label{eq:rdo}
  \text{min  }  \mathbb{E}  [d^*(\mathbf{Y},\hat{\mathbf{Y}})], \text{   s.t.  }  I(\mathbf{Y};\hat{\mathbf{Y}})\leq R,
\end{split}
\end{equation}
where $I(\cdot;\cdot)$ denotes the mutual information and $R$ denotes the bit-rate constraint.
% Note that, as mentioned above, some information of signal $Y$ cannot be observed directly.
%
% The compressed representation $h$ can only be generated from the noisy observation $X$. %,
The estimated $\hat{Y}$ can only be generated from the noisy observation $X$.
%called the indirect source coding problem~\cite{kipnis2021rate}.
%
According to existing works on indirect source coding~\cite{dobrushin1962information,wolf1970transmission}, if the distortion metric $d^{*}\left( \cdot \right)$ is known, the best R-D tradeoff is attained with the estimation-then-compression strategy, namely to first estimate $\hat{\mathbf{Y}}$ from $X$ and then compress $\hat{\mathbf{Y}}$ according to the R-D tradeoff directly. However, such optimality in the ideal circumstance is intractable and impractical in the intelligent visual analytics system targeting real applications:
\begin{itemize}
	\item \textbf{Absence of the accurate definition of}  $d^*(\cdot)$.
	% There are too many $d_t(\cdot)$ in real cases and it is hard to estimate their relative importance relationship.
  The aggregation function $f(\cdot)$ can vary case by case, corresponding to the relative importance among different $d_t(\cdot)$, which may also have various forms at different times.
	%
	% For example, the importance of different tasks can vary in various instances at different times. 
	% \item 
	% \textbf{Difficulty in determining} $\mathbf{Y}$ \textbf{when building codecs}.
	% We face massive applications and tasks. Especially, smart cities give birth to emerging new applications every day.
	\item \textbf{Intractable estimation of} $P(\hat{Y}_1, \hat{Y}_2, \cdots, \hat{Y}_N)$. The optimization of such a complex system, involving accurately estimating the joint probability $P(\hat{Y}_1, \hat{Y}_2, \cdots, \hat{Y}_N)$ for efficient compression, is usually intractable.
\end{itemize}}

To make the problem tractable in the real-world cases, we narrow down the scenarios into a series of Neural Network~(NN)-based applications. We show that with knowledgeable neural networks pre-trained on visual data, we can approach a practical system that solves the relaxed form of indirect source coding problem in the analytics taxonomy.
% The task set coding problem turns to equivalently compact, effective, and general feature extraction in neural networks.

% \subsection{Compact, Effective, and General Feature Extraction in NN}
\subsection{Compression for Analytics with Knowledgeable Neural Networks}

\wh{
We assume that for a specific task $\{X,Y_t\}$, a neural network with $M$ layers has been trained to predict $\hat{Y}_t$ from $X$. For simplicity, we neglect $t$ in the following notation and refer to $Y$ as the label for each task. The Markov chain involving the processing of the neural network can be formulated as,
\begin{equation}
  Y \xrightarrow{P_{X|Y}} X \xrightarrow{f_1} h_1 \xrightarrow{f_2} h_2 \rightarrow \cdots \rightarrow h_{M-1}\xrightarrow{f_M} \hat{Y},
\end{equation}
where $f_i$ denotes the processing function of the $i$-th layer.
According to the Information Bottleneck theory of neural networks~\cite{tishby2015deep},
a well-trained network tends to reduce the mutual information between $X$ and $h_i$ as $i$ increases,
by dismissing the irrelevant parts of $X$ w.r.t. $Y$.
Meanwhile, it preserves the mutual information $I(Y;\hat{Y})$ for accurately estimating $\hat{Y}$.
The goal of the information compression for analytics can be naturally implemented by compressing the latent representation $h_i$ and decoding the compressed representation for estimating $\hat{Y}$.
Compared with the raw scheme in Sec.~\ref{sec:fomulate1} that searches for the universally optimal representation to support multiple tasks jointly,
the relaxed problem has a smaller search space and becomes tractable.

When we look into the relaxed indirect source coding problem in the above-mentioned NN form, as shown in Fig.~\ref{fig:nn-gen}, three aspects have to be investigated as follows,

\noindent \textbf{Information Entropy of Representations.} We first aim to compress the deep features towards its bit-rate lower bound, \textit{i.e.} the information entropy. We propose a general dimension reduction-based compression model to measure the entropy of each latent representation $h_i$ produced in a multi-layer NN-based processing pipeline. 
The compression scheme employs a parametric entropy model to accurately estimate its probability distribution and the related entropy.
We define the \textit{plateau bit-rate} $R_p$ for $h_i$ as,
\begin{equation}
  \begin{split}
    &  R_p = \inf \mathbb{E}_{y \sim P_Y}[-\log p(z)], \text{   s.t. } {\mathbb{E}_{y\sim P_Y}[d(y,\hat{y}')] \leq \mathbb{E}_{y\sim P_Y}[d(y,\hat{y})]}, \\
    & \text{where  } h_i = f_i \circ f_{i-1}, \circ \cdots \circ, f_1(x), \text{  } x\sim P_{X|Y}, \\
    & z  = E(h_i), \text{  } \hat{y}' = f_M \circ f_{M-1} \circ \cdots \circ f_{i+1} \circ D(z).
  \end{split}
\end{equation}
% where $z$ is transformed from $h_i$ with $z = E(h_i)$, and the representation decoded from $z$ is then processed by the rest NN layers to produce prediction $\hat{y}'$ with $\hat{y}' = f_M \circ f_{M-1} \circ \cdots \circ f_{i+1} \circ D(z)$.
%namely the minimal bit-rate needed to achieve the average distortion as if the latent representation $h_i$ has not been compressed.
% namely, a threshold bit-rate beyond which the distortion will change little.
$R_p$ refers to the threshold bit-rate that, if more bit-rate is allowed beyond $R_p$, the distortion $\mathbb{E}[d(y,\hat{y}')]$ will not be improved, but remain approximately the same at the \textit{plateau}.
An encoder $E$ and a decoder $D$ are optimized to transform $h_i$ for entropy estimation and reconstruction for further processing, respectively.
%
% Under the compression scheme, if we only handle one task, it is observed that, as $i$ increases, $R_s$ decreases.
We also show that different tasks have different $R_p$.

\noindent \textbf{Aggregated Compression of Multiple Representations.}
While each compressed latent representation can be used to support a specific task, in the multi-task compression circumstance, 
independently compressing the representation of each task inevitably leads to inefficiency due to the cross-task redundancy.
We further investigate the issue of the transform to aggregate the multiple representations into a unified one.
We observe that while aggregation compression can reduce redundancy and save bit-rates, a side effect comes, namely that the analytics performance might be interfered.
When involving feature representations for other tasks (\textit{e.g.}, $h^2, h^3, \cdots, h^N$ for tasks $Y_2, Y_3, \cdots, Y_N$, respectively), the additional information in $h^2, h^3, \cdots, h^N$ tend to be the noise from the perspective of the intrinsic signal $h^1$ in the Markov chain $Y_1 \rightarrow X \rightarrow h_1^1 \rightarrow h_2^1 \rightarrow \cdots \rightarrow h_{M-1}^1 \rightarrow \hat{Y}_1$.
We further provide an analysis on different ways of aggregation, to improve compression efficiency while avoiding such a side effect.

\noindent \textbf{Generalizability of the Representations to Unseen Tasks.} 
With the proper design, the aggregated latent representation contains the information of multiple tasks and can effectively support these tasks. 
However, as stated in the problem formulation, we hope to support a super-set of any given set of tasks, thus the aggregated latent representation is expected to be generalized to handle unseen tasks.
% %
Therefore, we further study the potential of the compressed feature representation to generalize to external unseen tasks.
% Such generalizability is affected by the tradeoff between the overall bit-rate of the aggregated latent representation and the given tasks used in aggregation.
% %
% We further study the characteristics of this tradeoff under the proposed aggregated compression framework.
% %
% We show that there is a sweet point for such tradeoff in a set of latent representations, \textit{i.e} the features extracted from a series of neural networks serving for multiple tasks.

\section{Proposed Method}

% In the following, we first introduce the proposed compression model for neural network~(NN) generated feature maps $h_i$ from the input image $X$, as shown in Fig.~\ref{fig:codebook}. With the proposed compression model, we estimate the information entropy of each representation, and further design the aggregated compression framework for multiple tasks.

% \subsection{Compression Model for NN-Generated Feature Maps}
\subsection{Codebook-Hyperprior Model for Deep Feature Compression}

\begin{figure}[t]
  \centering
  \includegraphics[width=0.9\linewidth]{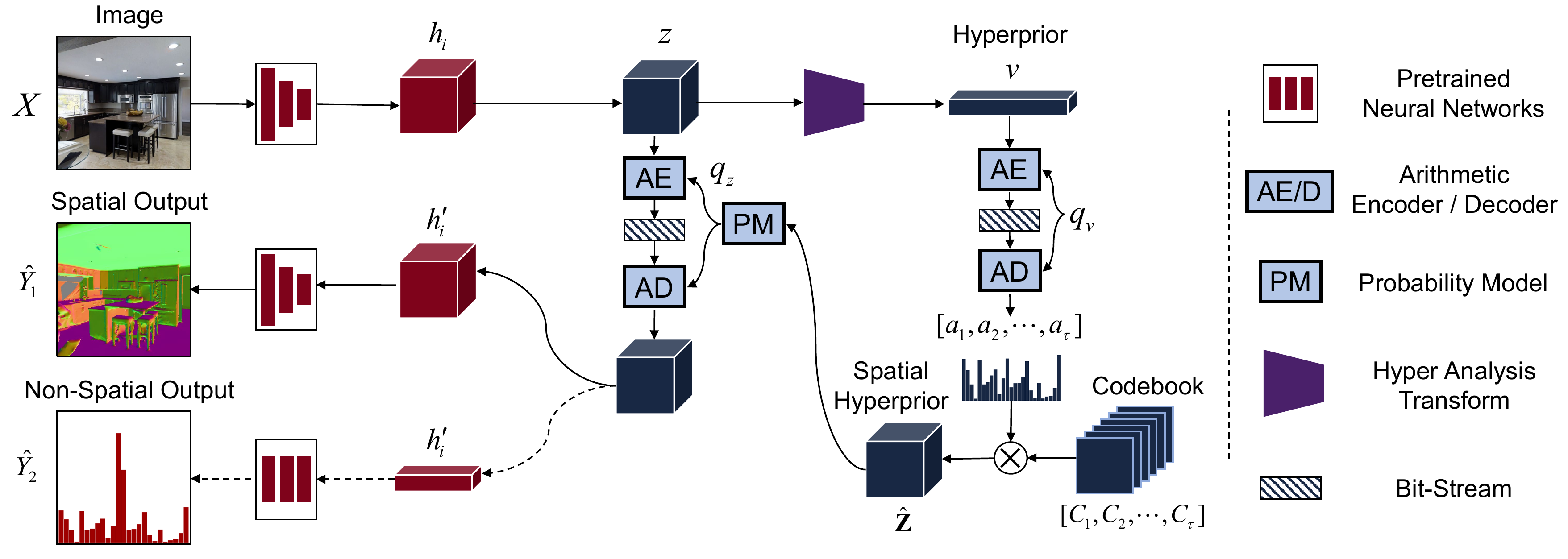}
  \caption{
    The codebook-hyperprior based data compression model for deep features. 
	With a pre-trained neural network (in \RRED{red}), a feature tensor is extracted and compressed by the proposed model (in \BLUE{blue}).
	The reconstructed feature tensor is processed by the rest layers of the pre-trained network to produce analytics results. Dashed lines illustrate the processing of the feature vectors without the spatial dimension. }
  \label{fig:codebook}
\end{figure}

In order to estimate the information entropy of each deep feature representation $h_i$ in the processing paradigm shown in Fig.~\ref{fig:nn-gen}, we design a compression model for the extracted deep features, illustrated in Fig.~\ref{fig:codebook}.
%
% Generally, there are two types of feature representations in neural networks, \textit{i.e.} feature tensors with spatial dimensions and features vectors without it. 
%
% These two kinds of representations serve pixel-wise analytics, \textit{e.g.} semantic parsing, and images-wise analytics, \textit{e.g.} attribute prediction, respectively. 
%
% We first take the feature tensors as an example, and later we show that the proposed model also supports feature vectors.
We aim to estimate the entropy of $h_i$ by compressing it to a bit-stream. 
As the mainstream neural networks do not apply any constraint on its generated $h_i$, the probability distribution of $h_i$ is usually unknown and it is intractable to estimate the entropy of $h_i$.
Therefore, we apply a transform to $h_i$ and obtain an equivalent representation $z$. The transform makes $z$ have the desired signal structure. Thus, its probability distribution is tractable.
Hence, we can estimate the entropy of $h_i$ via calculating the entropy of the structured representation $z$.
Specifically, $z$ has the following properties for easier entropy estimation.
Firstly, elements of $z$ have been quantized to integers. 
The value of each element $z_k$ belongs to a finite set $\mathbb{S}=\{t_{min}, \cdots, -1, 0, 1, \cdots, t_{max}\}$, 
and thus $z$ is sampled from the finite space $\mathbb{S}^K$, where $K$ denotes the dimension of $z$.
Given a probability distribution in the finite space, the information entropy can be calculated as,
\begin{equation}
  H(z) = \sum_{z \in \mathbb{S}^K} -p(z) \log p(z),
\end{equation}
which is the lower bound of the average bit-rate needed to encode $z$. 
As it is usually intractable to estimate $p_z$, we adopt a parametric probability model $q_z$ to estimate the probability distribution of $z$ during the encoding.
The actual bit-rate to encode $z$ with the probability $p_z$ under an estimated entropy model $q_z$ equals to the cross-entropy~\cite{cover1999elements} of $p$ and $q$, as,
\begin{equation}
  \begin{split}
    \label{eq:hpq}
    H(p, q) = \mathbb{E}_{p} [-\log q] = H(p) + D_{KL}(p||q).
  \end{split}
\end{equation}
It has been shown in Eq.~(\ref{eq:hpq}) that $H(p) \leq H(p,q)$, where the equality is achieved when $D_{KL}(p||q) = 0$, \textit{i.e.} when the probability model $q$ estimates $p$ perfectly.

Ball{\'e} \textit{et al.}~\cite{balle2018variational} propose to extract and encode a hyperprior from an image representation for more accurate entropy estimation control.
The hyperprior is used to estimate the probability distribution of the corresponding image representation, which is often a lower-resolution representation, and a hierarchical structure of a hyperprior can further improve the accuracy of the probability estimation for image representation~\cite{hu2020coarse}.
However, feature representations $h_i$ and $z$ are not image-level signals.
They are only expected to serve machine vision tasks and do not include the information of image appearances.
Although their extracted features might take the form of tensor and have the spatial dimensions, these features in fact are capable of being embedded into very low-dimensional space, which does not have the spatial dimensions.
As image compression oriented hyperprior model much relies on the assumption of hierarchical structure in images, it fails to capture the signal structure of $h_i$ and $z$, making the entropy estimation less effective.
% The strong assumption of resolutions hierarchy enforced by the image oriented hyperprior 
% Therefore, the previous image compression oriented hyperprior model faces the challenge. The hyperprior either has the spatial resolution, which fails to capture the signal structure of $h_i$ and $z$, or has no the spatial resolution, which makes the hyperprior model less effective.
It leads to a gap between $p$ and $q$.

To reduce the gap, we make the assumption that the extracted feature representations from the neural network can be embedded into a very low-dimensional manifold.
Each observed instance can be regarded as a point sampled from that the low-dimensional subspace, and the perturbation is independently distributed conditioned on the coordinates that expand the space.
This assumption naturally leads to the proposed low-dimensional hyperprior model.
The main idea is that, we adopt the hyperprior vector without the spatial dimensions in the encoding to capture the intrinsic signal structure of $h_i$ and $z$, but transform the hyperprior vector into the hyperprior tensor with the spatial dimensions in the decoding to augment the hyperprior's modeling capacity.
To estimate the probability distribution of $z$, a hyperprior $v$ is extracted from $z$ via a hyper analysis transform $f_{Ha}(\cdot)$ as, namely, $v = f_{Ha}(z)$.
% \begin{equation}
%   v = f_{Ha}(z).
% \end{equation}
The estimation of probability $p(z)$ can be divided into $p(z) = p(z,v) = p(v)p(z|v)$.
Then, we apply a global pooling operation to reduce the spatial dimensions of $z$, producing $v$ in the vector form. Note that $v$ is also quantized to integers.
We further assume that each element in $v$ follows a zero-mean Gaussian distribution $\mathcal{N}(0,\sigma_j)$, and conditioned on $v$, each element $z_k$ in $z$ is conditionally independently distributed. The entropy of $v$ is estimated by tuning the parameter $\sigma_j$.
%, therefore the probability $q_v$ is given by,
% \begin{equation}
%   \label{eq:qv}
%   q(v) = \prod_i^N \left[\phi\left(\frac{v_i+\frac{1}{2}}{\sigma_{i}}\right)-\phi\left(\frac{v_i-\frac{1}{2}}{\sigma_{i}}\right)\right],
% \end{equation}
% %
% where $\phi(\cdot)$ refers to the standard normal cumulative distribution function, and $\sigma_i$ is the parameter to be estimated in the training.

% We assume that
%
We model $q_{z_k|v}$ with a Gaussian distribution $q_{z_k|v} \sim \mathcal{N}(\mu_k=f(v;\theta_{f}), \sigma_k=g(v;\theta_{g}))$, where the mean and scale are generated through a function of $v$.
To achieve this, we decode $n$ sequences of coefficients from $v$.
Each sequence $\mathbf{A}^l = \left(a_1^l, a_2^l, \cdots, a_{\tau}^l\right), l \in [1,n]$ indicates a linear combination of the spatial bases, defined by a codebook, in the form of $\{C_1, C_2, \cdots, C_{\tau}\}$.
With the codebook and the sequences of coefficients $\{\mathbf{A}\}_n$, we generate the spatial hyperprior $\hat{\mathbf{Z}}$ as,
\begin{equation}
  \begin{split}
  \hat{Z}_l &= a_1^l C_1 + a_2^l C_2 + \cdots + a_{\tau}^l C_{\tau} \text{ , for  } l=1,2, \cdots, n,\\
  \hat{\mathbf{Z}} &= (\hat{Z}_1, \hat{Z}_2, \cdots \hat{Z}_n).
  \end{split}
\end{equation}
We employ a prediction sub-network to estimate $\mu_k=f(v;\theta_{f}), \sigma_k=g(v;\theta_{g})$ from $\hat{\mathbf{Z}}$.
By learning the parameters of the sub-network, $\theta_{f}$ and $\theta_{g}$ are estimated to provide an accurate estimation $q(z|v)$ for $p(z|v)$.
%
% Together with $q(v) \sim \mathcal{N}(0,\sigma_i)$, the entropy estimation is tractable with Eq.~(\ref{eq:hpq}).
%
The spatial dimensions of the codebook $\{C_1, C_2, \cdots, C_{\tau}\}$ are fixed, and therefore it requires a re-sampling to deal with the inputs of different resolutions.

The proposed model is also general and flexible to support the deep features without spatial dimensions,
\textit{i.e.} feature vectors. 
This can be achieved by directly producing the vector-form probability parameters $\mu_k=f(v;\theta_{f}), \sigma_k=g(v;\theta_{g})$ with $v$, via multi-layer perceptions.

\subsection{Aggregation Transformed Compression}

\begin{figure}[t]
  \centering
  \includegraphics[width=0.9\linewidth]{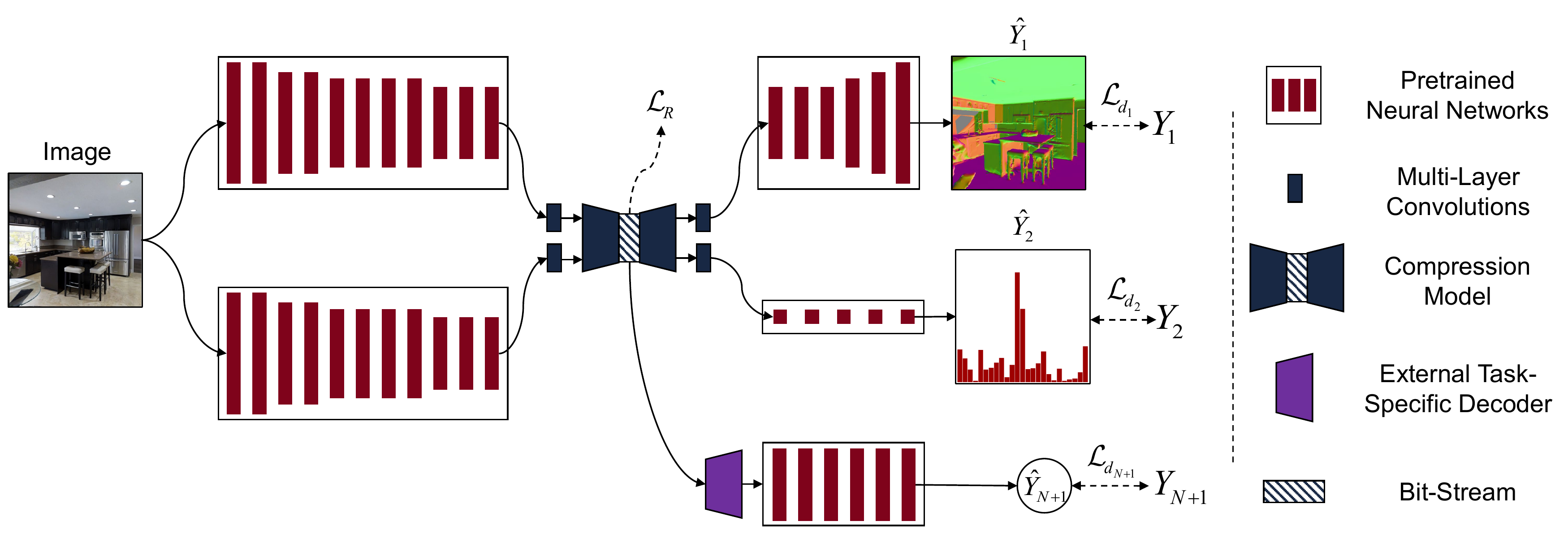}
  \caption{Aggregation transformed compression model for deep feature maps.}
  \label{fig:octopus}
  \vspace{-3mm}
\end{figure}

It has been shown in \cite{zamir2018taskonomy} that,
there exist connections among feature representations of different tasks.
Thus, if multiple tasks are supported as we mentioned in the problem formulation,
the separate compression for each task may be less efficient due to the cross-task redundancy.
Therefore, we propose the aggregation transformed compression scheme to generate the compressed representation for different tasks jointly.

An example of the proposed aggregation transformed compression scheme is shown in Fig.~\ref{fig:octopus}.
The illustrated structure compresses and aggregates the feature representations of two tasks into one bit-stream.
Each representation is transformed with a sub-network with convolutional layers. The transformed features are concatenated and compressed via a compression model.
The decompressed representation is then split via another set of convolutional layers, serving as the input of the rest of the pre-trained analytics network.

The aggregation transformed compression model is trained in two stages, 
corresponding to the two application scenarios, including 1) analytics oriented compression in a known set of tasks;
% \textit{i.e.} dealing with the tasks in a given set of tasks,
and 2) out-of-set analytics, \textit{i.e.} handling the unseen task.
During the first training phase, the parameters of the compression model 
and the multi-layer peripheral convolutions $f_{\text{perc}}\left( \cdot \right)$ before the compression model for each task are tuned.
Parameters of the pre-trained analytics models are fixed.
The compression model learns to compress different forms of feature representations jointly. The parameters are trained with the joint R-D loss function as,
\begin{equation}
  \begin{split}
    \mathcal{L} = \mathcal{L}_R + \lambda_1 \mathcal{L}_{d_1} + \lambda_2 \mathcal{L}_{d_2} + \cdots + \lambda_N \mathcal{L}_{d_N}, 
  \end{split}
\end{equation}
where $\lambda_i$ are Lagrange multipliers to indicate the relative importance of different tasks $i$, respectively.
The second training phase is triggered if an external unseen task is involved.
In this phase, the compression model is fixed to ensure that the compressed feature representation does not alter.
An external task-specific decoder is trained to decode the compressed feature representation $z$ to $\hat{h}_{N+1}$.
A task-related loss function $\mathcal{L}_{d_{N+1}}$ is applied to ensure that the decoded $\hat{h}_{N+1}$ can maximally utilize the information in the bit-stream to support the external task.}

	\section{Experiments}
	
	\subsection{Experimental Settings}
	% Dataset
	% which tasks
	
	We conduct the experiments on the Taskonomy dataset~\cite{zamir2018taskonomy}, which contains approximately 4.5 million images, all labeled by 25 attributes, to support various machine vision tasks. 
	The abundance of tasks link to one image provides the desired environment for our study. 
	We utilize the pre-trained models on the dataset, provided by the authors under the MIT License.
	All pre-trained models are hourglass encoder-decoder neural networks, as described in~\cite{zamir2018taskonomy}.
	
	The following experiments are conducted on a subset of the original raw data.
	The subsets are selected at random, while we control the numbers of images in the splits, \textit{i.e.} 51,316 images for training, 945 for validation, and 1,024 for testing. 
	Images in different splits of the data are captured in different buildings.
	Thus, the splits are diverse in content.
	We select a set of \textit{real-world} tasks for evaluation, \textit{i.e.} scene classification, object classification, semantic segmentation, surface normal estimation, reshading, and principle curvature estimation.
	The selected tasks include diversified categories, with which we evaluate both high-level semantics driven analytics and mid-level geometry related estimation. 
	
	\subsection{Efficacy of the Proposed Compression Scheme}

	\begin{table}[b]
    \footnotesize
		\centering
    \vspace{-5mm}
		\caption{Experimental results on the semantic segmentation task with various compression schemes. Method IDFC is evaluated with different QPs, marked as IDFC~(QP) in the table. $\uparrow$ means higher performance, better result, and $\downarrow$ vice versa.}
		\begin{tabular}{cccccc}
			\toprule
			Method & Bit-Rate (bpp)$\downarrow$ & Cross Entropy$\downarrow$ & Acc.$\uparrow$ & Non-BG Acc.$\uparrow$ & mIoU$\uparrow$ \\
			\midrule
			Original & /   & 0.74  & 91.64\% & 86.28\% & 27.65\% \\
			Control Group & /   & 0.61  & 92.31\% & 82.67\% & 27.07\% \\
			\midrule
			IDFC~(51)~\cite{chen2019toward} & 0.020 & 6.22  & 92.74\% & 15.62\% & 11.03\% \\
			IDFC~(43)~\cite{chen2019toward} & 0.026 & 1.38  & \textbf{93.94}\% & 72.24\% & 28.41\% \\
			Hyperprior~\cite{chamain2021end} & 0.025 & 0.80  & 91.95\% & 79.64\% & 25.42\% \\
			Ours  & \textbf{0.013} & \textbf{0.77}  & 93.58\% & \textbf{81.35}\% & \textbf{29.35}\% \\
			\bottomrule
		\end{tabular}%
		\label{tab:seg}%
    \vspace{-3mm}
	\end{table}%
	
	We first evaluate the efficacy of the proposed codebook-hyperprior driven compression model for deep feature representations.
	We compare our method with the intermediate deep feature compression~(IDFC) method~\cite{chen2019toward}, and the hyperprior model~\cite{balle2018variational} used in the feature compression scheme~\cite{chamain2021end}.
	Note that both the hyperprior model and the proposed scheme involve a training process.
	To avoid the potential bias due to the training procedure, we set up the \textit{Control Group} experiments, where a transform network with an identical structure to the compression model is trained, but no bit-rate constraint is applied. 
	For the evaluated semantic segmentation task, we train the models with the element-wise cross-entropy loss function, weighted by the parameters originally provided by \cite{zamir2018taskonomy}.
	We select the model checkpoint with the lowest R-D cost and compare on the testing set.
	Both hyperprior and the proposed model are trained with $\mathcal{L}=R+\lambda \mathcal{L}_{CE}$, where $\lambda=1$. 
	We follow the setting in \cite{zamir2018taskonomy} to compare on $256\times256$ images, and calculate bits-per-pixel on that resolution. 
	The results are shown in Table~\ref{tab:seg}, where \textit{Original} refers to the results given by the originally provided hourglass-like networks in \cite{zamir2018taskonomy}. 
	We calculate mean pixel-level accuracy~(Acc.), the accuracy of pixels in the non-background regions~(Non-BG Acc.), and mean IoU~(mIoU) by averaging the result among all 17 classes, respectively, to evaluate the semantic segmentation performance.
	
	% We refer to the \textit{plateau} bit-rate as that, at this bit-rate, the model can provide good enough analytics performance that is comparable to non-rate-control networks.
  % as if there is no rate control.
	%
	The range of bit-rates we show in Table~\ref{tab:seg} is regarded as the \textit{plateau} bit-rate, where the compressed feature representation provides enough information to make the prediction accuracy comparable to models without bit-rate control.
	It is suggested by the results that the proposed method can better compress the deep features than existing methods~\cite{chen2019toward,chamain2021end}, as it consumes less bit-rate to reach a higher analytics performance in multiple metrics. 
	
	\subsection{Plateau Bit-Rate in Different Tasks}
	
	\begin{table}[t]
    \footnotesize
		\centering
		\caption{Evaluation on the plateau bit-rate for different tasks with the proposed method and IDFC. We present the validation set performance (Val. Perf.) and the test set performance (Test Perf.) along with the related bit-rate. Performances of different tasks are evaluated in different metrics. $\uparrow$ means higher performance metric, better result, and $\downarrow$ vice versa.}
		\begin{tabular}{cccccc}
			\toprule
			Task  & Method & Val. Perf. & Val. bpp & Test Perf. & Test bpp \\
			\midrule
			\multirow{4}[0]{*}{Scene Class$\uparrow$ } & Original & 70.02\% & /     & 67.48\% & / \\
			& Control Group & 75.66\% & /     & 62.70\% & / \\
			& IDFC  & 61.16\% & 0.0403  & 65.43\% & 0.0408  \\
			& Ours  & 71.11\% & \textbf{0.0068}  & 59.47\% & \textbf{0.0069}  \\
			\midrule
			\multicolumn{1}{c}{\multirow{4}[0]{*}{Semantic Seg.$\uparrow$ }} & Original & 18.37\% & /     & 27.65\% & / \\
			& Control Group & 18.85\% & /     & 27.07\% & / \\
			& IDFC  & 17.20\% & 0.0210 & 28.41\% & 0.0261 \\
			& Ours  & 18.19\% & \textbf{0.0072} & 29.35\% & \textbf{0.0131} \\
			\midrule
			\multirow{4}[0]{*}{Surface Normal$\downarrow$} & Original & 0.0741 & /     & 0.1211 & / \\
			& Control Group & 0.0700 & /     & 0.1252 & / \\
			& IDFC  & 0.0753 & 0.0520 & 0.1281 & 0.0588 \\
			& Ours  & 0.0721 & \textbf{0.0187} & 0.1299 & \textbf{0.0197} \\
			\midrule
			\multirow{4}[0]{*}{Reshading$\downarrow$} & Original & 0.2209 & /     & 0.2836 & / \\
			& Control Group & 0.1687 & /     & 0.2343 & / \\
			& IDFC  & 0.2217 & 0.0830 & 0.2844 & 0.0959 \\
			& Ours  & 0.1713 & \textbf{0.0130} & 0.2411 & \textbf{0.0134} \\
			\bottomrule
		\end{tabular}%
		\label{tab:tasks}%
    \vspace{-4mm}
	\end{table}%
	
	With the proposed compression scheme, we study the plateau bit-rate \textit{w.r.t.} different tasks.
	In this experiment, we train compression models for each task, respectively, and measure the bit-rate of the compressed feature representations.
	We search for the minimal bit-rate needed to support a task to its maximally achievable performance by the provided feature, \textit{i.e.} to make the performance comparable to non-rate-control settings.
	The experiments involve four different tasks. 
	We measure the performance of each task in different criteria, \textit{i.e.} accuracy for scene classification (Scene Class), 
	mIoU for semantic segmentation (Semantic Seg.), and $L_1$ distance for surface normal estimation of indoor scenes (Surface Normal) as well as reshading of an indoor image (Reshading). 
	
	The results are shown in Table~\ref{tab:tasks}. 
	As shown, the performances of different tasks reach their plateau at different bit-rates, indicating that the information entropy to support a machine vision task varies among different tasks. 
    Image-level analytics, \textit{e.g.}, classification, requires less bit-rate to support, while pixel-level analytics require more. 
    There are also differences among pixel-level analytics.
    We also show that IDFC consumes significantly more bit-rates. 
    Besides, as IDFC involves a quantization based transform coding process, the quantization noise can result in unpredictable interference on the analytics performance. 
    The results suggest that such quantization noise degrades the analytics performance more significantly on the geometry related tasks. 
    Meanwhile, the proposed scheme provides better support for different kinds of tasks.
	
	\subsection{Aggregation Transform for Compression in Analytics Taxonomy}
	\label{sec:exp3}
	\begin{table}[t]
    \footnotesize
		\centering
		\caption{Analytics performance and the joint bit-rate \textit{w.r.t} different aggregation schemes. \textit{Customized} refers to independently compressing feature maps for each tasks. The \textit{Trinity} and \textit{Hex} settings are as described in the main text.}
		\begin{tabular}{cccc|ccc}
			\toprule
			Task  & Metric & Original & Control Group & Customized & Trinity & Hex \\
			\midrule
			Scene Class & Accuracy$\uparrow$ & 70.02\% & 75.74\% & \textbf{71.19\%} & 71.08\% & 62.18\% \\
			Semantic Seg, & mIoU$\uparrow$  & 18.37\% & 18.85\% & 18.19\% & 18.14\% & \textbf{20.30\%} \\
			Object Class & Accuracy$\uparrow$ & 60.17\% & 60.02\% & 61.55\% & \textbf{64.19\%} & 59.75\% \\
			Normal & $L_1$ Distance $\downarrow$  & 0.074 & 0.071 & \textbf{0.073} & \textbf{0.073} & 0.074 \\
			Reshading & $L_1$ Distance $\downarrow$ & 0.221 & 0.172 & 0.173 & \textbf{0.168} & \textbf{0.168} \\
			Curvature & $L_1$ Distance $\downarrow$ & 0.300 & 0.296 & \textbf{0.296} & 0.299 & 0.306 \\
			Total Bit-Rate &  Bpp Sum $\downarrow$ & /     & /     & 0.059 & \textbf{0.049} & 0.053 \\
			\bottomrule
		\end{tabular}%
		\label{tab:aggre}%
    \vspace{-4mm}
	\end{table}%
	
	In this experiment, we compare the aggregated transformed compression scheme with the customized compression setting for different tasks.
	We present the results on the validation set, shown in Table~\ref{tab:aggre}, where \textit{Hex} means jointly compress all six kinds of representations with the model in Fig.~\ref{fig:octopus}.
	We further investigate the \textit{Trinity} compression setting, by separating the six tasks into two groups, \textit{i.e.} \textbf{A}: Scene Class, Semantic Seg. and Object Class; \textbf{B}: Surface Normal, Reshading and Curvature.
	As shown in Table~\ref{tab:aggre}, the joint compression of multiple representations saves more bit-rate. When all tasks reach the performance plateau, the \textit{Trinity} setting saves about 16.9\% bit-rate than \textit{Customized} (the last row in Table~\ref{tab:aggre}).
	However, a larger aggregation set affects the analytics performance.
	This may be because the information from the external tasks tends to act as additional noise for the focused task.
	By grouping similar tasks in one aggregation, higher analytics performance and lower bit-rate can be achieved.
	
	\subsection{Supporting Unseen Tasks}
	
	% Table generated by Excel2LaTeX from sheet 'Sheet2'
	\begin{table}[t]
    \footnotesize
		\centering
		\caption{Evaluation of compression schemes to support unseen tasks at the plateau bit-rates.}
		\begin{tabular}{ccc|cc}
			\toprule
			Representation & bpp   & Object Class & bpp   & Reshading \\
			\midrule
			Original & /     & 60.17\% & /     & 0.221 \\
			Binary & \textbf{0.0132} & 51.06\% & 0.0229 & \textbf{0.194} \\
			Binary+ & 0.0137 & 53.50\% & \textbf{0.0167} & 0.205 \\
			BPG Image & 0.0371 & \textbf{54.56}\% & 0.0371 & 0.222  \\
			\bottomrule
		\end{tabular}%
		\label{tab:extern}%
    \vspace{-3mm}
	\end{table}%
	
	We further explore employing the compressive representation to support external tasks that are not used in R-D training. 
	We conduct the experiment in two trinity groups as described in Sec.~\ref{sec:exp3}, 
	while we train the compression model only for two supervision tasks.
	The representation is used to train an external decoder for an unseen task, as shown in Fig.~\ref{fig:octopus}. 
	When evaluating object classification, only scene classification and semantic segmentation are used for supervision. 
	The same goes for the reshading task, where only the surface normal and curvature tasks are used for supervision. 
	This is marked as \textit{Binary} representations in Table~\ref{tab:extern}. 
	We also note that in some application scenarios, although the compression component cannot be supervised by an unseen task, the pre-trained model for that task is available.
	Thus, in the \textit{Binary+} setting, the source feature for the third task is included in the compression but only the other two tasks are used for supervision. 
	We compare with BPG~\cite{bpg} compressed images, which is also task-independent. 
	The results on the validation set are shown in Table~\ref{tab:extern}.
	
	As shown, the proposed method can generate compressed visual representations that support external unseen tasks, 
	achieving better performance than utilizing image compression methods. 
	The results also indicate that including the external feature representation can further help improve the performance for the representation sensitive tasks, 
	\textit{e.g.}, object classification, although the R-D training is not supervised for that task.

\section{Conclusion}

In this paper, we formulate and study the problem of information compression in analytics taxonomy. We propose a codebook-hyperprior model for more efficient deep feature representation compression, with which we explore to analyze the information entropy of feature representations for a set of machine vision tasks. We further propose to jointly compress visual representations for different tasks, which saves the bit-rate and provides the support of external unseen tasks. With the study, we provide the insight in designing a more efficient remote visual data processing system.

\small
\bibliographystyle{ieee_fullname}
\bibliography{reference}

\appendix

% \section{Appendix}

% Optionally include extra information (complete proofs, additional experiments and plots) in the appendix.
% This section will often be part of the supplemental material.

\end{document}